\documentclass[10pt,twocolumn,letterpaper]{article}

\usepackage{cvpr}              

\usepackage[dvipsnames]{xcolor}

\definecolor{cvprblue}{rgb}{0.21,0.49,0.74}
\usepackage[pagebackref,breaklinks,colorlinks,citecolor=cvprblue]{hyperref}
\usepackage{pifont}

\newcommand{\cmark}{\ding{51}}
\DeclareRobustCommand{\myparagraph}[1]{\noindent\textbf{#1}} 

\usepackage{multirow}
\usepackage{newtxtext,newtxmath}

\newcommand*\colourcheck[1]{%
  \expandafter\newcommand\csname #1check\endcsname{\textcolor{ForestGreen}{\ding{52}}}%
}
\colourcheck{green}

\newcommand{\epic}{Epic-tent\xspace} 
\newcommand{\epicO}{Epic-tent-O\xspace} 
\newcommand{\assembly}{Assembly101\xspace} 
\newcommand{\assemblyO}{Assembly101-O\xspace} 

\usepackage[accsupp]{axessibility} 

\def\paperID{4850} 
\def\confName{CVPR}
\def\confYear{2024}

\title{PREGO: online mistake detection in PRocedural EGOcentric videos}

\author{Alessandro Flaborea*$^{\vardiamondsuit\clubsuit}$\space\space\space Guido Maria D'Amely di Melendugno*$^\vardiamondsuit$ \space\space\space Leonardo Plini$^\vardiamondsuit$ \space\space\space  Luca Scofano$^\vardiamondsuit$ \\ Edoardo De Matteis$^\vardiamondsuit$ \space\space\space Antonino Furnari$^\spadesuit$ \space\space\space Giovanni Maria Farinella$^{\ddag\spadesuit}$ \space\space\space Fabio Galasso$^{\ddag\vardiamondsuit}$ \\
{\tt\small \{flaborea,damely,dematteis,galasso\}@di.uniroma1.it \space\space \{plini,scofano\}@diag.uniroma1.it}\\
{\tt\small \{antonino.furnari,giovanni.farinella\}@unict.it } \\\\ $^{\vardiamondsuit}$Sapienza University of Rome, Italy  \space\space\space $^\clubsuit${ItalAI (italailabs.com)} \space\space\space $^{\spadesuit}$University of Catania, Italy}

\begin{document}
\maketitle
\def\thefootnote{*}\footnotetext{Authors contributed equally.}\def\thefootnote{\arabic{footnote}}
\def\thefootnote{\ddag}\footnotetext{Co-senior role.}\def\thefootnote{\arabic{footnote}}

\begin{abstract}
\label{sec:abstract}

Promptly identifying procedural errors from egocentric videos in an online setting
is highly challenging and valuable for detecting mistakes as soon as they happen. 
This capability has a wide range of applications across various fields, such as manufacturing and healthcare.
The nature of procedural mistakes is open-set since novel types of failures might occur, which calls for one-class classifiers trained on correctly executed procedures. 
However, no technique can currently detect open-set procedural mistakes online.
We propose PREGO, the first online one-class classification model for mistake detection in PRocedural EGOcentric videos. PREGO is based on an online action recognition component to model the current action, and a symbolic reasoning module to predict the next actions. Mistake detection is performed by comparing the recognized current action with the expected future one. We evaluate PREGO on two procedural egocentric video datasets, \assembly and \epic, which we adapt for online benchmarking of procedural mistake detection to establish suitable benchmarks, thus defining the \assemblyO and \epicO datasets, respectively. 
The code is available at \href{https://github.com/aleflabo/PREGO}{https://github.com/aleflabo/PREGO}.

\end{abstract}

\section{Introduction} \label{sec:intro}
\begin{figure}
    \centering
    \includegraphics
    [trim={3cm 0cm 3cm 0cm}, width=0.8\linewidth]
    {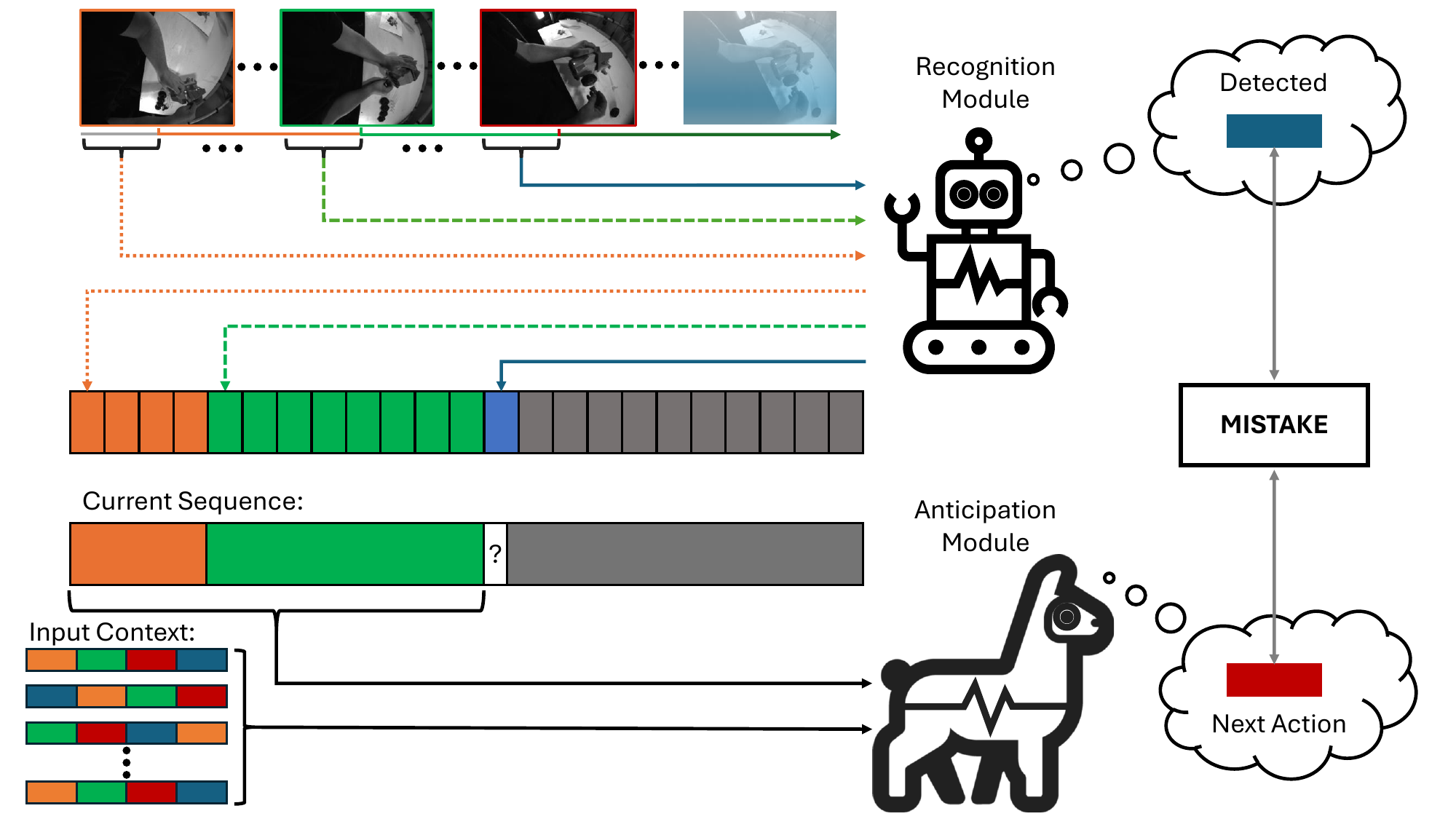}
    \caption{
     PREGO is based on two main components: The recognition module (top) processes the input video in an online fashion and predicts actions observed at each timestep; the anticipation module (bottom) reasons symbolically via a Large Language Model to predict the future action based on past action history and a brief context, such as instances of other action sequences. Mistakes are identified when the current action detected by the step recognition method differs from the one forecasted by the step anticipation module (right). 
     }
    \label{fig:conceptual}
\end{figure}

Egocentric procedure learning is gaining attention due to advancements in Robotics and Augmented Reality (AR) technologies. These technologies are pivotal to enhancing online\footnote{Most workflows can be aided by \emph{online} monitoring algorithms, which provide feedback to the operator in due course. However, they may lag due to processing or connectivity delays. We distinguish online from real-time, whereby the second has strict requirements of instantaneous response.} monitoring systems, offering real-time feedback, and improving operator efficiency in various fields. 
Recent works have produced numerous datasets~\cite{epicTent,IndustReal,sener2022assembly101,HoloAssist2023,Ghoddoosian_2023_ICCV,ding2023every,coin,ikea,IndustReal,cross_task,proceL_2019,howto100m_2019_miech,ragusa2023enigma51}, methodologies aimed at advancing procedure learning~\cite{POC_2022,Ghoddoosian_2023_ICCV,OadTR,zhong_2023,epicTent,coin} and error detection models~\cite{HoloAssist2023,ding2023every,IndustReal}.
Despite these advancements, as outlined in Table~\ref{tab:mistake_methods}, state-of-the-art methods typically focus on supervised and offline mistake detection.
They are unsuitable for situations requiring dynamic decision-making, specifically within an {\it online} setting, 
or when errors occur unpredictably, thus defining these instances as open-set conditions.

In this work, we propose the first model to detect PRocedural errors in EGOcentric videos (PREGO), which operates online, thus causal, and can recognize unseen procedural mistakes, fitting for open-set scenarios.
We prioritize egocentric videos due to their highly detailed perspective, essential for accurately identifying steps within procedures. 
Additionally, the widespread use of egocentric cameras in industries~\cite{Plizzari2023AnOI} necessitates the development of online error detection techniques to improve the safety and efficiency of workers.
The online attribute is achieved by analyzing input videos sequentially up to a given frame \(t\), ensuring that no future actions influence the current step recognition. 
On the other hand, open-set learning is performed by exclusively exposing PREGO to correct procedural sequences when predicting mistakes, following the One-Class Classification (OCC) paradigm \cite{zaheer2022generative, Flaborea_2023_ICCV}. Any step within a procedure that significantly diverges from the expected correct patterns is identified as an error, allowing PREGO to recognize a wide range of procedural mistakes without being confined to a restricted set of predefined ones. 

PREGO's architecture is dual-branched, as depicted in Fig.~\ref{fig:conceptual}.
The first branch, the {\it step-recognition branch}, analyzes frames in a procedural video up to a current time \(t\), aiming to classify the action being undertaken by the operator.
This branch can exploit the current state-of-the-art video-based online step recognition model, \cite{An2023MiniROADMR, OadTR}. 
Concurrently, the second branch is in charge of \emph{step-anticipation}, tasked to predict the action at time $t$, based solely on the steps up to $t-1$.
We propose using a pre-trained Large Language Model (LLM)~\cite{touvron2023llama} for zero-shot symbolic reasoning through contextual analysis~\cite{surismenon2023vipergpt,gupta2023visual,generalpatternmachines2023}. 
An error is detected upon a misalignment between the currently recognized action and the anticipated one, thereby signaling a deviation from the expected procedure.
Utilizing correctly executed procedures as instances in the query prompt obviates the necessity for additional model fine-tuning and leverages the pattern-completion abilities of LLMs.
Our proposed approach is an abstraction from the video content. Using labels allows for longer-term reasoning, as a label summarizes several frames. Also, this approach is an alternative to the carefully constructed action inter-dependency graphs~\cite{ashutosh2023video}. We demonstrate that symbolic reasoning subsumes understanding lengthy procedures and the action inter-dependencies, suggesting repositioning from semantic-based expressions of procedures to an implicit representation, where only patterns of symbols have to be recognized and predicted. By representing procedures as sequences and their steps as symbols, we let the predictor focus on the patterns that characterize the correct procedures. 

\begin{table*}[t]
\centering
\caption{Comparison among relevant models.
In the modalities column, \textit{RGB} stands for RGB images, \textit{H} for hand poses, \textit{E} for eye gaze, \textit{K} for keystep labels. Differently from previous works, we are the first to consider an egocentric one class and online approach to mistake detection. 
}\label{tab:mistake_methods}
\resizebox{1\textwidth}{!}{

\begin{tabular}{c|ccccccc}
\hline

\multicolumn{1}{c}{}                & \multicolumn{1}{|c}{\textbf{Ego}} &  \textbf{OCC} & \textbf{Online} & \textbf{Modalities} & \textbf{Task} & \multicolumn{1}{c}{\textbf{Datasets}} \\ \hline\hline

Ding et al. \cite{ding2023every} - \textit{Arxiv '23}  &  &  & & \textit{K} & Mistake Detection & \assembly  \cite{sener2022assembly101} \\

Wang et al. \cite{HoloAssist2023} - \textit{ICCV '23} & \cmark  &  & & \textit{RGB}+\textit{H}+\textit{E} & Mistake Detection & HoloAssist \cite{HoloAssist2023} \\

Ghoddoosian et al. \cite{Ghoddoosian_2023_ICCV} - \textit{ICCV '23} &  &  & & \textit{RGB} & Unknown sequence detection & ATA \cite{Ghoddoosian_2023_ICCV} and CSV \cite{Qian_2022_CVPR} \\ 

Schoonbeek et al. \cite{IndustReal} - \textit{WACV '24} & \cmark  &  & \cmark & Multi & Procedure Step Recognition &  IndustReal\cite{IndustReal} \\ \hline

\textbf{PREGO}  & \cmark  & \cmark & \cmark & \textit{RGB} & Mistake Detection & \textit{\assemblyO}, \textit{\epicO}                         \\ \hline

\end{tabular}}

\end{table*}


To support the evaluation of PREGO, we adapt the procedural benchmarks of \assembly~\cite{sener2022assembly101}  and \epic~\cite{epicTent}, formalizing
the novel task of online procedural mistake detection.
In the adapted online mistake detection
benchmarks, which we dub \assemblyO and \epicO, the model is tasked with detecting when a procedural mistake is made, thus compromising the procedure. 
The compromising mistake may be a wrong action or a relevant action performed in such an order that the action dependencies are not respected.

We summarize our contributions as follows:
\begin{itemize}
    \item We present PREGO, the first method designed for online and open-set detection of procedural errors in egocentric videos. PREGO's online feature ensures causal analysis by sequentially processing input videos up to a given frame, preventing future actions from influencing current step recognition.
    \item PREGO achieves open-setness by exclusively relying on correct procedural sequences at training time, following the One-Class Classification (OCC) paradigm. This allows PREGO to identify a wide range of procedural mistakes, avoiding confinement to a predefined set of errors and avoiding the need for fine-grained mistake annotations.
    \item We propose using a pre-trained LLM for zero-shot symbolic reasoning through contextual analysis to predict the next action.
    \item To evaluate PREGO, we introduce the novel task of online procedural mistake detection and re-arrange existing datasets to provide two new benchmarks, referred to as \assemblyO and \epicO.
\end{itemize}

\section{Related Work}
\label{sec:relworks}
 
\subsection{Procedural Mistake Detection}
Procedural learning has seen significant advancements with the creation of diverse datasets~\cite{coin,proceL_2019,cross_task,howto100m_2019_miech,ragusa2023enigma51} that provide insights into both structured~\cite{ikea,HoloAssist2023,Ragusa_meccano_2021} and unstructured~\cite{epicTent,epickitchens_Damen_2018} tasks, covering a spectrum from industrial assembly~\cite{Ragusa_meccano_2021,sener2022assembly101,IndustReal,ragusa2023enigma51} to daily cooking activities~\cite{epickitchens_Damen_2018,breakfasts,salads}. Despite the increased focus on this area, there is a notable lack of a unified methodology for mistake detection, resulting in fragmented literature and scarce evaluations.

{\bf Datasets.} ATA~\cite{Ghoddoosian_2023_ICCV} is a procedural dataset designed for offline mistake detection in assembling activities. It only reports video-level mistakes annotations, making it impractical for frame-based applications. \assembly~\cite{sener2022assembly101} is a large-scale video dataset that annotates frame-level mistakes. The videos represent actors assembling toys, and the dataset offers synchronized Ego-Exo views and hand-positions data. Another recent assembling dataset with frame-level annotations is IndustReal~\cite{IndustReal}. However, the authors consider a single toy, which results in a single procedure to be learned. \epic~\cite{epicTent} is a dataset with a different domain, as it reports actors building up a tent in an outdoor scenario. The participants have different degrees of expertise, and they naturally commit mistakes that have been annotated in \epic. Holoassist~\cite{HoloAssist2023} is a recent dataset that presents egocentric videos of people performing several manipulating tasks instructed by an expert. In this study, we employ~\cite{sener2022assembly101,epicTent} datasets since they give insights into errors happening during procedures in two different contexts, i.e., controlled industrial and outdoor environments\footnote{At the time of writing~\cite{HoloAssist2023,IndustReal} were unavailable publicly.}.

{\bf Methods.} In Table~\ref{tab:mistake_methods}, we report the main features of the recent approaches to Mistake Detection in procedural videos. In~\cite{Ghoddoosian_2023_ICCV}, the authors train an action recognizer model and consider error detection a semantic way of evaluating the segmentation results. Their method is thus explicitly offline, while PREGO aims to promptly detect procedure mistakes as soon as they occur. By contrast, \assembly~\cite{sener2022assembly101} and Holoassist~\cite{HoloAssist2023} apply the same error detection baselines on varying granularity but also operate offline, requiring video segmentation. Ding et al.~\cite{ding2023every} use knowledge graphs for error identification, bypassing video analysis and extracting procedural steps from transcripts, presenting a distinct methodology within the procedural learning field. PREGO diverges from these works as it leverages the video frames to detect the steps of the procedure online and leverages symbolic reasoning for an online assessment of the procedure's correctness. Moreover, acknowledging that the mistake detection task shares many aspects with the established field of Video Anomaly Detection, we design PREGO to work in an OCC framework. 
As motivated in~\cite{Flaborea_2023_ICCV,zaheer2022generative}, this choice ensures that PREGO is not constrained to detect only specific kinds of errors, as it is trained on sequences that do not contain mistakes.

\subsection{Steps recognition and anticipation}
Step recognition is the task of identifying actions within a procedure. Indeed, a procedure is an ordered sequence of steps that bring to the completion of a task.
Step recognition is crucial in areas such as autonomous robotics and educational technology. 
Recent contributions in this domain include \cite{Shah_2023_STEPs}, which uses a novel loss for self-supervised learning and a clustering algorithm to identify key steps in unlabeled procedural videos. 
\cite{POC_2022} introduces an action segmentation model using an attention-based structure with a Pairwise Ordering Consistency loss to learn the regular order of the steps in a procedure.
They devise a weakly supervised approach, using only the set of actions occurring in the procedure as labels, avoiding frame-level annotations. 
\cite{zhong_2023} approaches the task by leveraging online instructional videos to learn actions and sequences without manual annotations, blending step recognition with a deep probabilistic model to cater to step order and timing variability. 
Notably, An et al.~\cite{An2023MiniROADMR} proposed miniROAD explicitly targeting online action detection. They leverage an RNN architecture and regulate the importance of the losses during training to perform active action recognition.

On the other hand, step anticipation focuses on predicting forthcoming actions in a sequence crucial for real-time AI decision-making. 
\cite{Abdelsalam_2023_gepsan} addresses this by generating multiple potential natural language outcomes, pretraining on a text corpus to overcome the challenge of diverse future realizations. 
Additionally, the framework of~\cite{self_regulated} proposes solutions to future activity anticipation in egocentric videos, using contrastive loss to highlight novel information and a dynamic reweighing mechanism to focus on informative past content, thereby enhancing video representation for accurate future activity prediction.
Unlike prior works, PREGO is the first model that anticipates actions via LLM symbolic reasoning in the label space. 

\subsection{Large Language Modelling and Symbolic Reasoning}

LLMs are trained on large datasets and have many parameters, giving them novel capabilities compared to previous language models~\cite{Wei2022EmergentAO}.
LLMs have shown remarkable abilities in modeling many natural language-related~\cite{touvron2023llama} and unrelated tasks~\cite{brooks2023instructpix2pix, Wei2022EmergentAO, gupta2023visual}.
Their next-token prediction mechanism aligns with our action anticipation branch, where both systems aim to infer future actions based on collected data.

Recent research~\cite{Pallagani2022PlansformerGS, gupta2023visual, Liang2022CodeAP, Feng2023LanguageMC, generalpatternmachines2023} has explored LLMs' ability to operate as \textit{In-Context Learners} (ICLs), which means they can solve novel and unseen tasks.
Given a query prompt with a context of input-output examples, LLMs can comprehend and address the problems in this setting without further fine-tuning.
LLMs as ICLs have been used for a variety of tasks, including planning~\cite{Pallagani2022PlansformerGS}, programming~\cite{gupta2023visual, Liang2022CodeAP}, logical solvers~\cite{Feng2023LanguageMC}, and symbolic reasoning~\cite{generalpatternmachines2023}. 

Some work has shown that LLMs can generate semantically significant patterns~\cite{generalpatternmachines2023}, while~\cite{Wei2023LargerLM} has explored LLMs' in-context capabilities on semantically unrelated labels, where there is no relationship between a token and its meaning. 
Recent works~\cite{Pallagani2022PlansformerGS,olmo2021gpt3} studied the opportunity to employ LLMs for devising plans to accomplish tasks. 
In our mistake detection pipeline, we leverage ICL using an LLM as our action anticipation branch. 
Given examples of similar procedures, such LLM continues sequences of steps in a procedure, represented as symbols.
The LLM acts as a symbolic pattern machine, continuing the pattern of actions given a context of sequences performed goal-oriented, even if the sequences do not follow a semantic scheme. 
This combines the challenges of predicting future actions and of having no semantics.

\section{Benchmarking online open-set procedural mistakes}
\label{sec:benchmarking}

This section presents the benchmark datasets and the evaluation metrics used in our experiments. 
First, we introduce the reviewed online variants of \assembly and  \epic (Sec.~\ref{sec4:datasets}), and then we define the proposed online metrics in Sec.~\ref{sec4:metrics}. 

\subsection{Datasets}
\label{sec4:datasets}
We propose {\it \assemblyO} and {\it \epicO} as a refactoring of the original datasets~\cite{sener2022assembly101,epicTent}, detailing the selected labeling for online benchmarking, and the novel arrangement of training and test splits, to account for open-set procedural mistakes.

\subsubsection{\assemblyO}
\label{sec4.1:Assembly101}
\assembly~\cite{sener2022assembly101} is a large-scale video dataset that enables the study of procedural video understanding. The dataset consists of 362 procedures of people performing assembly and disassembly tasks on 101 different types of toy vehicles. Each procedure is recorded from static (8) and egocentric (4) cameras and annotated with multiple levels of granularity, such as more than 100K coarse and 1M fine-grained action segments and 18M 3D hand poses. The dataset covers various challenges, including action anticipation and segmentation, mistake detection, and 3D pose-based action recognition.

\myparagraph{\assembly for online and open-set mistake detection (\textit{Proposed})}
We introduce a novel split of the dataset~\cite{sener2022assembly101} that enables online, open-set mistake detection by design. \assemblyO mainly encompasses two edits on~\cite{sener2022assembly101}, namely, a new train/test split and a revision of the length of the procedures. The novel split encloses all the correct procedures in the train set, leaving the 
videos with mistakes for the test and validation set. This modification is needed to allow models to learn the sequences of steps that characterize correct procedures in a one-class classification fashion. In this way, models do not undergo the bias of learning specific kinds of mistakes during training; instead, as they are exposed exclusively to correct processes, they adhere to the OCC protocol and consider mistakes all actions that diverge from the learned normalcy. As a further advantage, this saves all mistaken annotated videos for the test set, granting better balanced correct/mistaken validation and test sets and a more comprehensive evaluation of mistake detection. The second revision involves evaluating each video for benchmarking until the procedure is compromised, meaning until a mistake occurs due to incorrect action dependencies. Indeed, coherently with the OCC protocol, models are tasked with learning the correct flows of steps that allow procedures to be efficiently completed and considering sub-process after a mistake occurs creates a gap between the actions in the train set and those in the test, which prevents the models from recognizing or correctly anticipating the procedure steps. Moreover, this work proposes to focus on egocentric videos to be consistent with real-world applications. Hence, we only leverage a single egocentric video from the four views available for each video in~\cite{sener2022assembly101}. 

\subsubsection{\epicO}

 \epic is a dataset of egocentric videos that capture the assembly of a camping tent outdoors. The dataset was collected from 24 participants who wore two head-mounted cameras (GoPro and SMI eye tracker) while performing the task. The dataset contains 5.4 hours of video recordings and provides annotations for the action labels, the task errors, the self-rated uncertainty, and the gaze position of the participants. The dataset also reflects the variability and complexity of the task, as the participants interacted with non-rigid objects (such as the tent, the guylines, the instructions, and the tent bag) and exhibited different levels of proficiency and uncertainty in completing the task.

\myparagraph{ \epic for online and open-set mistake detection (\textit{Proposed})}
This section introduces a novel split for the  \epic dataset~\cite{epicTent}, designed to be adapted for the open-set mistake detection task. It is labeled with nine distinct mistake types. However, among these, ``{\it slow}", ``{\it search}", ``{\it misuse}", ``{\it motor}", and ``{\it failure}" do not represent procedural errors, since, when they occur, the procedure is not tainted. On the other hand, the categories ``{\it order}", ``{\it omit}", ``{\it correction}", and ``{\it repeat}" are procedural mistakes, which we consider for our task.
\begin{figure}[t]
    \centering
    \includegraphics
    [trim={3cm 2cm 3cm 0cm}, clip,
    width=\linewidth]{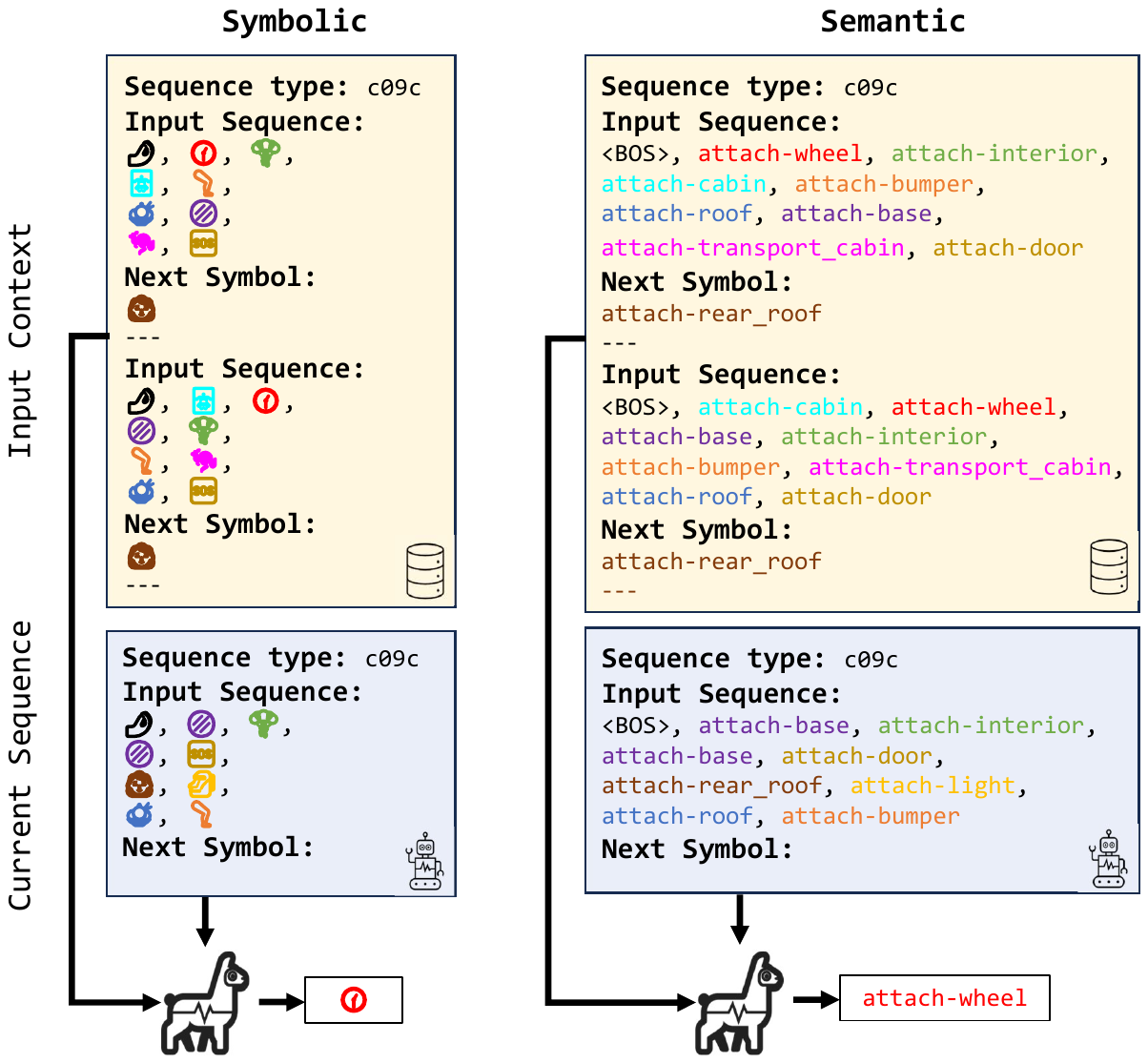}
    \caption{
    Two different representations of the actions in the prompt for the LLM model. On the left, the prompt is represented using symbolic labels. On the right, the prompt encompasses the names of the actions in the transcript. The context part of the prompt is fixed and retrieved from the dataset, while the recognition module extracts the current sequence.
    }
    \label{fig:frame_pic}
\end{figure}
 \epic is designed for the supervised error detection task and, differently from~\cite{sener2022assembly101}, every reported procedure includes some mistakes, hampering the reproduction of the split procedure proposed for \assemblyO. Nonetheless, this dataset provides the confidence scores assigned to each frame by the performer, indicating their self-assessed uncertainty during the task. Thus, we define a strategy for splitting , reported in Sec. C of the supplementary materials, 
 in which videos featuring the most confident performers form the train set, while those showing higher uncertainty (and thus potentially more prone to errors) populate the test set. This partitioning strategy holds encouraging promise, especially in real-world scenarios where the accurate labeling of erroneous frames is hard to achieve or where the training of a mistake detector can initiate immediately post-recording without necessitating the completion of the entire annotation process. The resulting split comprises 14 videos for the training set and 15 for the test set. 

The \epic dataset showcases only egocentric videos recorded through Go-Pro cameras. This further highlights the practicality and relevance of the proposed novel benchmark in open-scene contexts. The videos in the test set are also trimmed up to the last frame of the first mistake occurring in the video, while those representing correct procedures are maintained unaltered.

\subsection{Metrics}
\label{sec4:metrics}
To assess the performance of our procedural mistake detection model, we use True Positives as a measure of the model correctly identifying errors and True Negatives as a measure of accurately labeling steps that are not errors.
Thus, we rely on the Precision, Recall, and F1 score metrics to evaluate the performance of our model.
These metrics offer valuable insights into the model's capability to identify and classify mistakes within procedural sequences.
More specifically, precision quantifies the accuracy when predicting mistakes, minimizing false positives. Recall assesses the model's capability to retrieve all mistakes, reducing the number of false negatives. Finally, the F1 score is the harmonic mean of precision and recall, and it balances failures due to missing mistakes and reporting false alarms.

\section{Methodology}
\label{sec:methodology} 

PREGO exploits a dual-branch architecture that integrates procedural step recognition with anticipation modeling, as depicted in Fig.~\ref{fig:conceptual}.
In the following sections, we elaborate on the problem formalization (Sec. \ref{sec3:prego}), present the branches for step-recognition (Sec. \ref{sec3:oadtr}) and step-anticipation (Sec. \ref{sec3:symb}), and finally we illustrate the mistake detection procedure (Sec. \ref{sec3:mistake}).

\subsection{Problem Formalization}
\label{sec3:prego}
We consider a finite set of $N$ procedures $\{p_i\}_{i=1}^N$ 
that encodes the sequence of actions as $p_i = \{ a_k\}_{k=1}^{K_i}$ where $K$ varies depending on the specific procedure $i$ and $a_k \in \mathcal{A}= \{a | a \text{ is a possible action}\}$.
Each procedure is also represented by a set of videos 
that are composed of frames $v_i=\{f_\tau\}_{\tau=1}^{M_i}$ where $M_i$ is the total number of frames in the video $i$.
\\Fixed a frame $f_{\tau}$ from a given video $v_{i}$, PREGO's task is double-folded: it has to (1) recognize the action $a_{\tau}$ corresponding to the frame $f_{\tau}$ in the video and (2) predict the action $a_{\tau}$ that will take place 
at time $\tau$ considering only past observations until time $\tau-1$.
\\The step recognition task is performed by a module $\rho$ that takes as input the encoded frames of $v_{i}$ up to $\tau$ and returns an action $a^{\rho}_{\tau}$. 
We then feed the module $\xi$, responsible for the anticipation task, with all the $a^{\rho}_1, ...,a^{\rho}_{\tau-1} $ actions to have a prediction $a^{\xi}_{\tau}$ for the next action in the obtained sequence.
\\Finally, we compare $a^{\rho}_{\tau}$ with $a^{\xi}_{\tau}$
and we deem as mistaken the actions 
where a misalignment between the outputs of the two branches occurs. For clarity, in the remainder of this section, we consider a single procedure \(p\) associated with a video \(v\).

\subsection{Step Recognition}
\label{sec3:oadtr}
The step recognition module, denoted as $\rho$, receives encoded frames from $v_{i}$ up to $\tau$ as input and generates the action $a^{\rho}_{\tau}$. This module can be designed in a modular fashion under the condition that the model operates online, meaning it lacks knowledge of future events. In our approach, we leverage MiniRoad~\cite{An2023MiniROADMR}, renowned for its state-of-the-art performance in online action detection, its efficiency in computational complexity (measured in GFlops), and parameter count.

Within this framework, with $w$ representing the size of window $W$, the model forecasts the action $a_{\tau}$ by considering frames ${f_{{\tau}-w},..,f_{\tau}}$. However, this approach yields redundant outcomes as the model frequently predicts the same action for consecutive frames. We adopt a simple procedure to ensure consistency: we only consider unique actions whenever the model predicts the same action for consecutive frames.
The loss for this step recognition module is calculated through a Cross Entropy Loss, comparing the actual action $a_{\tau}$ with the predicted action $a^{\rho}_{\tau}$.

\subsection{Step Anticipation}
\label{sec3:symb}
We introduce a novel approach for step forecasting in procedural learning by harnessing the power of symbolic reasoning~\cite{generalpatternmachines2023} via a Language Model (LM). Specifically, we employ a Large Language Model (LLM) as our $\xi$ model for next-step prediction, feeding it with prompts from procedural video transcripts. These prompts are structured in two parts: the first part comprises contextual transcripts \(C\), \textit{Input Context} in Fig.~\ref{fig:frame_pic}, extracted from similar procedures to inform the LLM about typical step sequences and order. 
The second part, \textit{Sequence} in the Figure, includes the current sequence of actions up to a specific frame, \(f_{\tau}\), detected by our module $\rho$, i.e., 
\begin{equation}
s_{\tau} = [a^{\rho}_{1},...,a^{\rho}_{\tau-1}]  
\end{equation}
This approach enables the LLM to utilize in-context learning, eliciting its ability to anticipate subsequent actions. Our framework operates in a zero-shot fashion, relying on the LLM's ability to retrieve the correct sequence continuation without specific training or fine-tuning but only leveraging the positive examples within the input prompts. Additionally, our method employs symbolic representations of the steps, converting the set of actions $\mathcal{A}$ into a symbolic alphabet $\Omega$ through an invertible mapping $\gamma$. Therefore, we can express the symbolic predicted sequence as:
\begin{equation}
\gamma(s_{\tau})=[\gamma(a^{\rho}_{1}), ..., \gamma(a^{\rho}_{\tau-1})]=[\omega_1, ..., \omega_{\tau-1}]
\end{equation}
This conversion abstracts the actions from their semantic content, allowing the LLM to focus on pure symbols and sequences, thus simplifying the complexity of predicting the following action. \\ 
Finally, the \(\xi\) module, given the examples \(C\) and the current symbolic transcript \(\gamma(s_{\tau})\) described in its prompt, is required to output the most probable symbol $\omega_{\tau}$ to continue the sequence (see Figure~\ref{fig:frame_pic}).
At this point, we apply the inverse function of $\gamma$ to retrieve the underlying step label, i.e.,  \(a^{\xi}_{\tau} = \gamma^{-1}( \omega_{\tau})\).

\subsection{Mistake Detection}\label{sec4:PMD_formal}
We finally compare the outputs of the two modules to detect procedural mistakes. Precisely, we consider as correct all the steps where the outputs of the two modules align with each other, while we deem as an error the cases for which the two outputs diverge. That is:
\begin{equation}
\begin{cases}
    a^{\rho}_{\tau} \neq a^{\xi}_{\tau} & \text{MISTAKE} \\
    a^{\rho}_{\tau} = a^{\xi}_{\tau} & \text{CORRECT} \\
\end{cases}
\end{equation}
\label{sec3:mistake}

\section{Experiments}
\label{sec:experiments}

\begin{table*}[t]

\centering{
    \caption{
        A comparative assessment between PREGO and the chosen baseline methods is conducted to detect procedural mistakes using the \assemblyO and \epicO datasets. 
    }
    \label{tab:together}
    \resizebox{0.85\linewidth}{!}{
        \begin{tabular}{l|c|c|ccc|ccc} 
        \hline
         & \multicolumn{2}{c|}{\textbf{}} & \multicolumn{3}{c|}{\textbf{\assemblyO}}  & \multicolumn{3}{c}{\textbf{\epicO}} \\ 
         & \multicolumn{1}{c|}{\textbf{Step Recog.}} & \multicolumn{1}{c|}{\textbf{Step Antic.}} &  \multicolumn{1}{c}{\textbf{Precision}} & \multicolumn{1}{c}{\textbf{Recall}}  & \multicolumn{1}{c|}{\textbf{F1 score}} & \multicolumn{1}{c}{\textbf{Precision}} & \multicolumn{1}{c}{\textbf{Recall}} & \multicolumn{1}{c}{\textbf{F1 score}}\\ 
        \hline\hline
        \multirow{1}{*}{One-step memory} &  \textit{Oracle} &  & 16.3 & 30.7 & 21.3 & 6.6 & 26.6 & 10.6\\
        \multirow{1}{*}{BERT \cite{Devlin2019BERTPO}} &  \textit{Oracle} &  & 78.2  & 20.0  & 31.8 & 75.0 & 5.6 & 10.4\\
        \multirow{1}{*}{\textit{PREGO}} & \textit{Oracle} &   \textit{GPT-3.5}  & 29.2 & 75.8 & 42.1 & 9.9 & 73.3 & 17.4 \\

        \multirow{1}{*}{\textit{PREGO}} & \textit{Oracle}&  \textit{LLAMA} & 30.7 & 94.0 & 46.3 & 10.7 & 86.7 & 19.1 \\
        
        \hline
        \multirow{1}{*}{OadTR for MD~\cite{OadTR}}  &  \textit{OadTR~\cite{OadTR}} &  \textit{OadTR~\cite{OadTR}} & 24.3  & 18.1 & 20.7   & 6.7 & 21.7 & 10.2\\
        \multirow{1}{*}{\textit{PREGO}} &   \textit{OadTR~\cite{OadTR}} &  
         \textit{LLAMA} & 22.1 & \textbf{94.2} & 35.8 & 9.5 & \textbf{93.3} & \textbf{17.2} \\

        \multirow {1}{*}{\textit{PREGO}} &  \textit{MiniRoad~\cite{An2023MiniROADMR}} &  \textit{GPT-3.5} & 16.2 & 87.5 & 27.3 & 4.3 & 66.6 & 8.0 \\ 
        \multirow{1}{*}{\textit{PREGO}} &  \textit{MiniRoad~\cite{An2023MiniROADMR}} & 
        \textit{LLAMA} & \textbf{27.8} & 84.1 & \textbf{41.8} & \textbf{8.6} & 20.0 & 12.0 \\

        \hline

        \end{tabular}
    }
}
\vspace{-0.3cm}
\end{table*}
In this section, we present the results of our experiments on online and open-set mistake detection in procedural videos. We contrast PREGO with several baselines that employ different mistake detector techniques or use the ground truth as an oracle. The oracular scenario represents an upper bound for a given anticipation method since the recognition branch does entirely rely on the ground truth. 
All the baselines are assessed on the \assemblyO and \epicO datasets, detailed in section \ref{sec4:datasets}. Evaluation metrics include precision, recall, and F1 score, as outlined in \ref{sec4:metrics}. Baselines are introduced in section \ref{sec:baselines}, and the primary results are analyzed in \ref{sec:results}. Furthermore, we explore the influence of different prompt types in \ref{sec:symbolic} and the context in \ref{sec:context}. Lastly, implementation specifics are discussed in \ref{sec:implementation_det}, along with addressing certain limitations.

\subsection{Baselines}
\label{sec:baselines}

\noindent To estimate the effectiveness of PREGO, we evaluate its performance  by comparing it against the following baseline models based on the metrics presented in Sec.~\ref{sec4:metrics}:

\myparagraph{One-step memory} We define a {\it transition matrix} considering only the correct procedures. Specifically, given the set of the actions  $\mathcal{A}$ in the training set with $|\mathcal{A}|=C$, we define a transition matrix $M\in\mathbb{R}^{C\times C}$ 
which stores in position $(l,m)$ the occurences that action $m$ follows action $l$. 
We then label as {\it mistake} the actions occurring in the test split that do not correspond to transitions recorded in the training set.

\myparagraph{OadTR for mistake detection}
The work~\cite{OadTR} proposes a framework for online action detection called OadTR that employs a Vision Transformer to capture the temporal structure and context of the video clips. The framework consists of an encoder-decoder architecture. The encoder takes the historical observations as input and outputs a task token representing the current action. The decoder takes the encoder output and the anticipated future clips as input and outputs a refined task token incorporating the future context.
In the context of procedural error detection, a mistake is identified when the output from the encoder does not align with the one from the decoder.

\myparagraph{BERT~\cite{Devlin2019BERTPO}}
We leverage the capability of BERT utilizing its specific [CLS] token to predict the correct or erroneous sequence of action. More specifically, we fine-tune BERT using the next-sentence-prediction task, where the model is trained to predict whether one sentence logically follows another within a given text. In our context, we apply this to determine whether step B can follow another step A within a procedure. Here, steps are defined as sets of two words, such as \textit{attach wheel}, representing coarse actions. To perform this, BERT is presented with pairs of sentences corresponding to actions A and B, tasking it with predicting the sequential relationship between them. BERT's advantage lies in pre-training on a vast text corpus, followed by fine-tuning for our specific scenario. This process enables BERT to grasp contextual connections between sentences, rendering it effective for tasks like classifying procedures and comprehending the logical flow of information in text.

\subsection{Results}
\label{sec:results}
We evaluate PREGO's performance on two datasets, \assemblyO and \epicO, and detail the results in Table~\ref{tab:together}. 
We replaced the step recognition branch's predictions with ground truth action labels to assess the upper bound on performance without step detection bias defining the {\it Oracle} setting.
This approach simulates a scenario where the video branch perfectly recognizes actions in the videos.
The One-step memory method considers only the previous action, while BERT reasons at a higher level of abstraction and leverages past actions more effectively. This reduces false alarms but introduces a conservative bias in the form of missing mistakes.
PREGO outperformed all baselines by leveraging symbolic reasoning for richer context modeling. PREGO$_{LLama}$ achieved the highest F1-score with a 45.6\% improvement over BERT, 
demonstrating the effectiveness of symbolic reasoning. Among PREGO configurations, PREGO$_{LLAMA}$ performed 9\% better than PREGO$_{GPT-3.5}$ on \assemblyO, due to its more powerful symbolic representation. Similar trends are observed on \epicO with metric values influenced by dataset characteristics (\epicO allows for more diverse assembly procedures compared to \assemblyO).

We move beyond oracle methods that rely on ground truth information and compare PREGO's performance against the established method OadTR~\cite{OadTR} per-frame action detection and forecasting. PREGO$_{Llama}$, using the same method for step recognition, significantly outperforms OadTR for MD achieving a 102\% improvement in F1-score (refer to Table~\ref{tab:together} for detailed results). 
OadTR is restricted to processing fixed-size video segments with a default window of 64 frames, resulting in the smallest F1-score. Indeed, it is insufficient for capturing the context of long procedures lasting an average of 7 minutes in \assembly.
The improvement can also be attributed to PREGO's symbolic step anticipation branch. Symbolic reasoning allows PREGO to operate at a higher level of abstraction than video-based methods like OadTR. This advantage mitigates video-based approaches' challenges with occlusion and forecasting fine-grained actions.

PREGO$_{LLama}$ can better learn the normal patterns of the procedures and detect deviations from them, achieving the best results in terms of F1-score.
In addition, PREGO$_{GPT-3.5}$ incurs costs that scale with the number of processed tokens, hindering its suitability for large-scale studies. LLAMA, being open-source, facilitates cost-effective exploration of PREGO at scale.
Compared to their oracle counterparts, PREGO$_{GPT-3.5}$ and PREGO$_{LLama}$ could potentially gain 54\% and 11\% improvement in F1-score, respectively. This suggests that the video branch's accuracy bottlenecks overall performance. However, the oracle recognition experiment also highlights the potential for improvement within PREGO itself. Other factors influencing performance include the quality of symbolic inputs, semantic prompts, and the underlying LLM architecture.

\subsection{Performance of Different Prompt Types}
\label{sec:symbolic}
We investigate the effect of different action representations in the prompt for the Step Anticipation task. Following \cite{generalpatternmachines2023}, we consider three ways of representing an action: numerical, semantic, or random symbols. Numerical representation means that an action label is replaced with an index in the range $[0,\mathcal{A}]$, where $\mathcal{A}$ is the total number of actions. Semantic representation implies that the action is represented by its action label. Random symbol indicates that each action is assigned to a different symbol, such as a set of emojis. This allows us to examine how the LLM can manage different levels of abstraction and expressiveness of the input prompt. Fig. \ref{fig:frame_pic} illustrates an example of the same prompt in two representations, symbolic and semantic. 

Table \ref{tab:symbolic} shows the experiment results using the described representations. 
We observe that all the different representations achieve close performance, with the random representation achieving the highest F1 score, 41.8, followed by the semantic and numerical representations, with 41.4 and 39.9, respectively.
We hypothesize that employing a numerical system to represent different actions might inadvertently introduce a form of bias related to ordering. This type of bias occurs because the relationship between specific actions and their corresponding numerical values is inherently arbitrary, lacking a natural or logical sequence. As a result, the numerical mapping can obscure the characteristics of the actions being represented, leading to potential challenges in accurately anticipating or predicting future actions based on these numerical representations.
Remarkably, the semantic representation achieves a comparable performance even though words can introduce bias or ambiguity into the model. This indicates that PREGO can handle the natural language input and extract the relevant information for the step anticipation task. 
Surprisingly, the random symbol representation has the highest performance amongst the other representations, even though the model has no semantic or numerical association with them. This suggests that the model effectively learns the temporal structure of the actions from the input history, regardless of the symbol representation. 

\subsection{Performance of Different Prompt Context}
\label{sec:context}
We examine two alternative ways of writing a prompt (Table~\ref{tab:symbolic_ablation}) for the PREGO method: prompting with a less representative context Vs.\ a more elaborate one.
The less informative prompt, labeled as ``Unreferenced-Context" in Table, requests PREGO to produce the next step without providing the model with the information that the contexts are sequences and that the output required is a symbol. The context is simply given as ``Context", the current sequence is given as ``Input", and the next step is requested as ``Output". 
The more elaborate prompt, labeled ``Elaborate" in Table, has a more complex prompt for both the context and the output. The context is given with the sentence ``Given the sequences of the following type:", the sequence to be completed as ``Complete the following sequence", and the output ``Sequence is completed with". The three prompts are shown in Fig. 1 of the supplementary materials.

The results show that the referenced-context prompt achieves the best F1 score (41.8). The other two alternatives perform similarly, reaching an F1 score of 41.4 and 40.5.
The detailed prompt structure is the most effective way of writing a prompt for the PREGO method, as it clearly conveys the essential information and the objective of the task. 

\begin{table}[t]

\centering{

\caption{
Performance of PREGO with different prompt representations for Procedural Mistake Detection evaluated via F1 score, precision and recall on the \assemblyO dataset.
}
\label{tab:symbolic}
\resizebox{0.8\linewidth}{!}{
\begin{tabular}{lccc} 
\hline
 & \multicolumn{1}{c}{\textbf{Precision}} & \multicolumn{1}{c}{\textbf{Recall}} & \multicolumn{1}{c}{\textbf{F1 score}}  \\ 
\hline\hline

\multirow{1}{*}{Numerical} & 26.7 & 78.6  & 39.9 \\
\multirow{1}{*}{Semantic}  & {\bf 27.8} & 81.3 & 41.4 \\
\multirow{1}{*}{Random}  & \textbf{27.8} & \textbf{84.1} & \textbf{41.8} \\

\hline
\end{tabular}
}
}

\end{table}

\subsection{Implementation Details}
\label{sec:implementation_det}
PREGO is trained on two P6000 GPUs using the Adam optimizer, a batch size of 128, a learning rate of \(1e^{-5}\), and a weight decay of \(1e^{-4}\). 
For Assembly-101-O, we use the pre-extracted TSN frame level features from \cite{sener2022assembly101}. For Epic-Tent-O, we extract the features using the same method. The training process takes approximately 4 hours.
PREGO achieves 0.02 fps on an NVIDIA Quadro P6000, meeting our needs without real-time constraints. 

\myparagraph{Limitations}
Across the currently available procedural datasets with annotated mistakes, the number of procedures only ranges up to hundreds, which is a limitation for current deep learning techniques. The original \assembly~\cite{sener2022assembly101} dataset encompasses 330 procedures; our proposed \assemblyO inherits only the procedures without mistakes as the learning set, namely 190 procedures; similarly, both \epic~\cite{epicTent} and \epicO only include 29 videos depicting the same task. We acknowledge the need for a large-scale dataset for online mistake detection and leave it as a future work. Indeed, more procedures will likely let the models generalize better, improving their capability to deal with multiple plausible procedures.
\begin{table}[t]

\centering{

\caption{
Impact of prompt variations on PREGO - Unreferenced-Context, Elaborate, and Referenced-Context prompts. Evaluated via F1 score, precision and recall on the \assemblyO dataset.
}
\label{tab:symbolic_ablation}
\resizebox{\linewidth}{!}{
\begin{tabular}{lccc} 
\hline
 & \multicolumn{1}{c}{\textbf{Precision}} & \multicolumn{1}{c}{\textbf{Recall}}  & \textbf{F1 score} \\ 
\hline\hline
\multirow{1}{*}{Elaborate} & 26.9 & 82.4 & 40.5 \\
\multirow{1}{*}{Unreferenced-Context} & 27.3 & 85.2 & 41.4 \\
\multirow{1}{*}{Referenced-Context (\textit{PREGO})} & \textbf{27.8} & \textbf{84.1} & \textbf{41.8} \\

\hline
\end{tabular}
}
}

\end{table}

\section{Conclusion}
\label{sec:conclusion}
We have introduced PREGO, a one-class, online approach for mistake detection in procedural egocentric video. PREGO predicts mistakes by comparing the current action predicted by an online step recognition model with the next action, anticipated through symbolic reasoning performed via LLMs. For evaluation, we adapt two datasets of procedural egocentric videos for the proposed task, thus defining the \assemblyO and \epicO datasets. Comparisons against different baselines show the feasibility of the proposed approach to one-class online mistake detection. We hope that our investigation and the proposed benchmark and model will support future research in this field.

\section*{Acknowledgements}

\noindent This work was carried out while L. Plini was enrolled in the Italian National Doctorate on Artificial Intelligence run by Sapienza University of Rome. We thank DsTech S.r.l. and the PNRR MUR project PE0000013-FAIR (CUP: B53C22003980006 and E63C22001940006) for partially funding the Sapienza University of Rome and University of Catania. 

\label{sec:acknowledgements}

{
    \small
    \bibliographystyle{ieeenat_fullname}
    \bibliography{main}

\begin{thebibliography}{41}
\providecommand{\natexlab}[1]{#1}
\providecommand{\url}[1]{\texttt{#1}}
\expandafter\ifx\csname urlstyle\endcsname\relax
  \providecommand{\doi}[1]{doi: #1}\else
  \providecommand{\doi}{doi: \begingroup \urlstyle{rm}\Url}\fi

\bibitem[Abdelsalam et~al.(2023)Abdelsalam, Rangrej, Hadji, Dvornik, Derpanis, and Fazly]{Abdelsalam_2023_gepsan}
Mohamed~A. Abdelsalam, Samrudhdhi~B. Rangrej, Isma Hadji, Nikita Dvornik, Konstantinos~G. Derpanis, and Afsaneh Fazly.
\newblock Gepsan: Generative procedure step anticipation in cooking videos.
\newblock In \emph{Proceedings of the IEEE/CVF International Conference on Computer Vision (ICCV)}, pages 2988--2997, 2023.

\bibitem[An et~al.(2023)An, Kang, Han, Yang, and Kim]{An2023MiniROADMR}
Joungbin An, Hyolim Kang, Su~Ho Han, Ming-Hsuan Yang, and Seon~Joo Kim.
\newblock Miniroad: Minimal rnn framework for online action detection.
\newblock \emph{2023 IEEE/CVF International Conference on Computer Vision (ICCV)}, pages 10307--10316, 2023.

\bibitem[Ashutosh et~al.(2023)Ashutosh, Ramakrishnan, Afouras, and Grauman]{ashutosh2023video}
Kumar Ashutosh, Santhosh~Kumar Ramakrishnan, Triantafyllos Afouras, and Kristen Grauman.
\newblock Video-mined task graphs for keystep recognition in instructional videos.
\newblock \emph{arXiv preprint arXiv:2307.08763}, 2023.

\bibitem[Ben-Shabat et~al.(2021)Ben-Shabat, Yu, Saleh, Campbell, Rodriguez-Opazo, Li, and Gould]{ikea}
Yizhak Ben-Shabat, Xin Yu, Fatemeh Saleh, Dylan Campbell, Cristian Rodriguez-Opazo, Hongdong Li, and Stephen Gould.
\newblock The ikea asm dataset: Understanding people assembling furniture through actions, objects and pose.
\newblock In \emph{Proceedings of the IEEE/CVF Winter Conference on Applications of Computer Vision}, pages 847--859, 2021.

\bibitem[Brooks et~al.(2023)Brooks, Holynski, and Efros]{brooks2023instructpix2pix}
Tim Brooks, Aleksander Holynski, and Alexei~A Efros.
\newblock Instructpix2pix: Learning to follow image editing instructions.
\newblock In \emph{Proceedings of the IEEE/CVF Conference on Computer Vision and Pattern Recognition}, pages 18392--18402, 2023.

\bibitem[Damen et~al.(2018)Damen, Doughty, Farinella, Fidler, Furnari, Kazakos, Moltisanti, Munro, Perrett, Price, and Wray]{epickitchens_Damen_2018}
Dima Damen, Hazel Doughty, Giovanni~Maria Farinella, Sanja Fidler, Antonino Furnari, Evangelos Kazakos, Davide Moltisanti, Jonathan Munro, Toby Perrett, Will Price, and Michael Wray.
\newblock Scaling egocentric vision: The epic-kitchens dataset.
\newblock In \emph{European Conference on Computer Vision (ECCV)}, 2018.

\bibitem[Devlin et~al.(2019)Devlin, Chang, Lee, and Toutanova]{Devlin2019BERTPO}
Jacob Devlin, Ming-Wei Chang, Kenton Lee, and Kristina Toutanova.
\newblock Bert: Pre-training of deep bidirectional transformers for language understanding.
\newblock In \emph{North American Chapter of the Association for Computational Linguistics}, 2019.

\bibitem[Ding et~al.(2023)Ding, Sener, Ma, and Yao]{ding2023every}
Guodong Ding, Fadime Sener, Shugao Ma, and Angela Yao.
\newblock Every mistake counts in assembly.
\newblock \emph{arXiv preprint arXiv:2307.16453}, 2023.

\bibitem[Elhamifar and Naing(2019)]{proceL_2019}
Ehsan Elhamifar and Zwe Naing.
\newblock Unsupervised procedure learning via joint dynamic summarization.
\newblock 2019.

\bibitem[Feng et~al.(2023)Feng, Xu, Hao, Sharma, Shen, Zhao, and Chen]{Feng2023LanguageMC}
Jiazhan Feng, Ruochen Xu, Junheng Hao, Hiteshi Sharma, Yelong Shen, Dongyan Zhao, and Weizhu Chen.
\newblock Language models can be logical solvers.
\newblock \emph{ArXiv}, abs/2311.06158, 2023.

\bibitem[Flaborea et~al.(2023)Flaborea, Collorone, di~Melendugno, D'Arrigo, Prenkaj, and Galasso]{Flaborea_2023_ICCV}
Alessandro Flaborea, Luca Collorone, Guido Maria~D'Amely di Melendugno, Stefano D'Arrigo, Bardh Prenkaj, and Fabio Galasso.
\newblock Multimodal motion conditioned diffusion model for skeleton-based video anomaly detection.
\newblock In \emph{Proceedings of the IEEE/CVF International Conference on Computer Vision (ICCV)}, pages 10318--10329, 2023.

\bibitem[Ghoddoosian et~al.(2023)Ghoddoosian, Dwivedi, Agarwal, and Dariush]{Ghoddoosian_2023_ICCV}
Reza Ghoddoosian, Isht Dwivedi, Nakul Agarwal, and Behzad Dariush.
\newblock Weakly-supervised action segmentation and unseen error detection in anomalous instructional videos.
\newblock In \emph{Proceedings of the IEEE/CVF International Conference on Computer Vision (ICCV)}, pages 10128--10138, 2023.

\bibitem[Gupta and Kembhavi(2023)]{gupta2023visual}
Tanmay Gupta and Aniruddha Kembhavi.
\newblock Visual programming: Compositional visual reasoning without training.
\newblock In \emph{Proceedings of the IEEE/CVF Conference on Computer Vision and Pattern Recognition}, pages 14953--14962, 2023.

\bibitem[Jang et~al.(2019)Jang, Sullivan, Ludwig, Gilchrist, Damen, and Mayol-Cuevas]{epicTent}
Youngkyoon Jang, Brian Sullivan, Casimir Ludwig, Iain Gilchrist, Dima Damen, and Walterio Mayol-Cuevas.
\newblock Epic-tent: An egocentric video dataset for camping tent assembly.
\newblock In \emph{Int. Conf. Comput. Vis.}, 2019.

\bibitem[Kuehne et~al.(2014)Kuehne, Arslan, and Serre]{breakfasts}
H. Kuehne, A.~B. Arslan, and T. Serre.
\newblock The language of actions: Recovering the syntax and semantics of goal-directed human activities.
\newblock In \emph{Proceedings of Computer Vision and Pattern Recognition Conference (CVPR)}, 2014.

\bibitem[Liang et~al.(2022)Liang, Huang, Xia, Xu, Hausman, Ichter, Florence, and Zeng]{Liang2022CodeAP}
Jacky Liang, Wenlong Huang, F. Xia, Peng Xu, Karol Hausman, Brian Ichter, Peter~R. Florence, and Andy Zeng.
\newblock Code as policies: Language model programs for embodied control.
\newblock \emph{2023 IEEE International Conference on Robotics and Automation (ICRA)}, pages 9493--9500, 2022.

\bibitem[Lu and Elhamifar(2022)]{POC_2022}
Zijia Lu and Ehsan Elhamifar.
\newblock Set-supervised action learning in procedural task videos via pairwise order consistency.
\newblock In \emph{IEEE Conf. Comput. Vis. Pattern Recog.}, pages 19903--19913, 2022.

\bibitem[Miech et~al.(2019)Miech, Zhukov, Alayrac, Tapaswi, Laptev, and Sivic]{howto100m_2019_miech}
Antoine Miech, Dimitri Zhukov, Jean-Baptiste Alayrac, Makarand Tapaswi, Ivan Laptev, and Josef Sivic.
\newblock How{T}o100{M}: {L}earning a {T}ext-{V}ideo {E}mbedding by {W}atching {H}undred {M}illion {N}arrated {V}ideo {C}lips.
\newblock In \emph{ICCV}, 2019.

\bibitem[Mirchandani et~al.(2023)Mirchandani, Xia, Florence, Ichter, Driess, Arenas, Rao, Sadigh, and Zeng]{generalpatternmachines2023}
Suvir Mirchandani, Fei Xia, Pete Florence, Brian Ichter, Danny Driess, Montserrat~Gonzalez Arenas, Kanishka Rao, Dorsa Sadigh, and Andy Zeng.
\newblock Large language models as general pattern machines.
\newblock In \emph{Proceedings of the 7th Conference on Robot Learning (CoRL)}, 2023.

\bibitem[Olmo et~al.(2021)Olmo, Sreedharan, and Kambhampati]{olmo2021gpt3}
Alberto Olmo, Sarath Sreedharan, and Subbarao Kambhampati.
\newblock Gpt3-to-plan: Extracting plans from text using gpt-3.
\newblock \emph{FinPlan 2021}, page~24, 2021.

\bibitem[Ouyang et~al.(2022)Ouyang, Wu, Jiang, Almeida, Wainwright, Mishkin, Zhang, Agarwal, Slama, Ray, et~al.]{ouyang2022training}
Long Ouyang, Jeffrey Wu, Xu Jiang, Diogo Almeida, Carroll Wainwright, Pamela Mishkin, Chong Zhang, Sandhini Agarwal, Katarina Slama, Alex Ray, et~al.
\newblock Training language models to follow instructions with human feedback.
\newblock \emph{Advances in Neural Information Processing Systems}, 35:\penalty0 27730--27744, 2022.

\bibitem[Pallagani et~al.(2022)Pallagani, Muppasani, Murugesan, Rossi, Horesh, Srivastava, Fabiano, and Loreggia]{Pallagani2022PlansformerGS}
Vishal Pallagani, Bharath Muppasani, Keerthiram Murugesan, Francesca Rossi, L. Horesh, Biplav Srivastava, F. Fabiano, and Andrea Loreggia.
\newblock Plansformer: Generating symbolic plans using transformers.
\newblock \emph{ArXiv}, abs/2212.08681, 2022.

\bibitem[Plizzari et~al.(2023)Plizzari, Goletto, Furnari, Bansal, Ragusa, Farinella, Damen, and Tommasi]{Plizzari2023AnOI}
Chiara Plizzari, Gabriele Goletto, Antonino Furnari, Siddhant Bansal, Francesco Ragusa, Giovanni~Maria Farinella, Dima Damen, and Tatiana Tommasi.
\newblock An outlook into the future of egocentric vision.
\newblock \emph{ArXiv}, abs/2308.07123, 2023.

\bibitem[Qi et~al.(2023)Qi, Wang, Su, Su, Huang, and Tian]{self_regulated}
Zhaobo Qi, Shuhui Wang, Chi Su, Li Su, Qingming Huang, and Qi Tian.
\newblock Self-regulated learning for egocentric video activity anticipation.
\newblock \emph{IEEE Transactions on Pattern Analysis and Machine Intelligence}, 45\penalty0 (6):\penalty0 6715--6730, 2023.

\bibitem[Qian et~al.(2022)Qian, Luo, Lian, Tang, Zhao, and Gao]{Qian_2022_CVPR}
Yicheng Qian, Weixin Luo, Dongze Lian, Xu Tang, Peilin Zhao, and Shenghua Gao.
\newblock Svip: Sequence verification for procedures in videos.
\newblock In \emph{IEEE Conf. Comput. Vis. Pattern Recog.}, pages 19890--19902, 2022.

\bibitem[Ragusa et~al.(2021)Ragusa, Furnari, Livatino, and Farinella]{Ragusa_meccano_2021}
Francesco Ragusa, Antonino Furnari, Salvatore Livatino, and Giovanni~Maria Farinella.
\newblock The meccano dataset: Understanding human-object interactions from egocentric videos in an industrial-like domain.
\newblock In \emph{Proceedings of the IEEE/CVF Winter Conference on Applications of Computer Vision (WACV)}, pages 1569--1578, 2021.

\bibitem[Ragusa et~al.(2024)Ragusa, Leonardi, Mazzamuto, Bonanno, Scavo, Furnari, and Farinella]{ragusa2023enigma51}
Francesco Ragusa, Rosario Leonardi, Michele Mazzamuto, Claudia Bonanno, Rosario Scavo, Antonino Furnari, and Giovanni~Maria Farinella.
\newblock Enigma-51: Towards a fine-grained understanding of human-object interactions in industrial scenarios.
\newblock In \emph{Proceedings of the IEEE/CVF Winter Conference on Applications of Computer Vision (WACV)}, 2024.

\bibitem[Schoonbeek et~al.(2024)Schoonbeek, Houben, Onvlee, van~der Sommen, et~al.]{IndustReal}
Tim~J Schoonbeek, Tim Houben, Hans Onvlee, Fons van~der Sommen, et~al.
\newblock Industreal: A dataset for procedure step recognition handling execution errors in egocentric videos in an industrial-like setting.
\newblock In \emph{Proceedings of the IEEE/CVF Winter Conference on Applications of Computer Vision}, pages 4365--4374, 2024.

\bibitem[Sener et~al.(2022)Sener, Chatterjee, Shelepov, He, Singhania, Wang, and Yao]{sener2022assembly101}
F. Sener, D. Chatterjee, D. Shelepov, K. He, D. Singhania, R. Wang, and A. Yao.
\newblock Assembly101: A large-scale multi-view video dataset for understanding procedural activities.
\newblock \emph{IEEE Conf. Comput. Vis. Pattern Recog.}, 2022.

\bibitem[Shah et~al.(2023)Shah, Lundell, Sawhney, and Chellappa]{Shah_2023_STEPs}
Anshul Shah, Benjamin Lundell, Harpreet Sawhney, and Rama Chellappa.
\newblock Steps: Self-supervised key step extraction and localization from unlabeled procedural videos.
\newblock In \emph{Proceedings of the IEEE/CVF International Conference on Computer Vision (ICCV)}, pages 10375--10387, 2023.

\bibitem[Stein and McKenna(2013)]{salads}
Sebastian Stein and Stephen~J. McKenna.
\newblock Combining embedded accelerometers with computer vision for recognizing food preparation activities.
\newblock In \emph{Proceedings of the 2013 ACM International Joint Conference on Pervasive and Ubiquitous Computing}. Association for Computing Machinery, 2013.

\bibitem[Sur\'is et~al.(2023)Sur\'is, Menon, and Vondrick]{surismenon2023vipergpt}
D\'idac Sur\'is, Sachit Menon, and Carl Vondrick.
\newblock Vipergpt: Visual inference via python execution for reasoning.
\newblock \emph{Proceedings of IEEE International Conference on Computer Vision (ICCV)}, 2023.

\bibitem[Tang et~al.(2019)Tang, Ding, Rao, Zheng, Zhang, Zhao, Lu, and Zhou]{coin}
Yansong Tang, Dajun Ding, Yongming Rao, Yu Zheng, Danyang Zhang, Lili Zhao, Jiwen Lu, and Jie Zhou.
\newblock Coin: A large-scale dataset for comprehensive instructional video analysis.
\newblock \emph{2019 IEEE/CVF Conference on Computer Vision and Pattern Recognition (CVPR)}, pages 1207--1216, 2019.

\bibitem[Touvron et~al.(2023)Touvron, Martin, Stone, Albert, Almahairi, Babaei, Bashlykov, Batra, Bhargava, Bhosale, et~al.]{touvron2023llama}
Hugo Touvron, Louis Martin, Kevin Stone, Peter Albert, Amjad Almahairi, Yasmine Babaei, Nikolay Bashlykov, Soumya Batra, Prajjwal Bhargava, Shruti Bhosale, et~al.
\newblock Llama 2: Open foundation and fine-tuned chat models.
\newblock \emph{arXiv preprint arXiv:2307.09288}, 2023.

\bibitem[Wang et~al.(2021)Wang, Zhang, Qing, Shao, Zuo, Gao, and Sang]{OadTR}
Xiang Wang, Shiwei Zhang, Zhiwu Qing, Yuanjie Shao, Zhengrong Zuo, Changxin Gao, and Nong Sang.
\newblock Oadtr: Online action detection with transformers.
\newblock In \emph{Int. Conf. Comput. Vis.}, pages 7565--7575, 2021.

\bibitem[Wang et~al.(2023)Wang, Kwon, Rad, Pan, Chakraborty, Andrist, Bohus, Feniello, Tekin, Frujeri, Joshi, and Pollefeys]{HoloAssist2023}
Xin Wang, Taein Kwon, Mahdi Rad, Bowen Pan, Ishani Chakraborty, Sean Andrist, Dan Bohus, Ashley Feniello, Bugra Tekin, Felipe~Vieira Frujeri, Neel Joshi, and Marc Pollefeys.
\newblock Holoassist: an egocentric human interaction dataset for interactive ai assistants in the real world.
\newblock In \emph{Int. Conf. Comput. Vis.}, pages 20270--20281, 2023.

\bibitem[Wei et~al.(2022)Wei, Tay, Bommasani, Raffel, Zoph, Borgeaud, Yogatama, Bosma, Zhou, Metzler, hsin Chi, Hashimoto, Vinyals, Liang, Dean, and Fedus]{Wei2022EmergentAO}
Jason Wei, Yi Tay, Rishi Bommasani, Colin Raffel, Barret Zoph, Sebastian Borgeaud, Dani Yogatama, Maarten Bosma, Denny Zhou, Donald Metzler, Ed~Huai hsin Chi, Tatsunori Hashimoto, Oriol Vinyals, Percy Liang, Jeff Dean, and William Fedus.
\newblock Emergent abilities of large language models.
\newblock \emph{Trans. Mach. Learn. Res.}, 2022, 2022.

\bibitem[Wei et~al.(2023)Wei, Wei, Tay, Tran, Webson, Lu, Chen, Liu, Huang, Zhou, and Ma]{Wei2023LargerLM}
Jerry~W. Wei, Jason Wei, Yi Tay, Dustin Tran, Albert Webson, Yifeng Lu, Xinyun Chen, Hanxiao Liu, Da Huang, Denny Zhou, and Tengyu Ma.
\newblock Larger language models do in-context learning differently.
\newblock \emph{ArXiv}, abs/2303.03846, 2023.

\bibitem[Zaheer et~al.(2022)Zaheer, Mahmood, Khan, Segu, Yu, and Lee]{zaheer2022generative}
M~Zaigham Zaheer, Arif Mahmood, M~Haris Khan, Mattia Segu, Fisher Yu, and Seung-Ik Lee.
\newblock Generative cooperative learning for unsupervised video anomaly detection.
\newblock In \emph{Proceedings of the IEEE/CVF conference on computer vision and pattern recognition}, pages 14744--14754, 2022.

\bibitem[Zhong et~al.(2023)Zhong, Yu, Bai, Li, Yan, and Li]{zhong_2023}
Y. Zhong, L. Yu, Y. Bai, S. Li, X. Yan, and Y. Li.
\newblock Learning procedure-aware video representation from instructional videos and their narrations.
\newblock In \emph{2023 IEEE/CVF Conference on Computer Vision and Pattern Recognition (CVPR)}. IEEE Computer Society, 2023.

\bibitem[Zhukov et~al.(2019)Zhukov, Alayrac, Cinbis, Fouhey, Laptev, and Sivic]{cross_task}
Dimitri Zhukov, Jean-Baptiste Alayrac, Ramazan~Gokberk Cinbis, David Fouhey, Ivan Laptev, and Josef Sivic.
\newblock Cross-task weakly supervised learning from instructional videos.
\newblock In \emph{CVPR}, 2019.

\end{thebibliography}
}


\clearpage

\twocolumn[
\begin{@twocolumnfalse}
\begin{center}
  \LARGE \textbf{Supplementary Materials}
\end{center}
\end{@twocolumnfalse}
]

This section provides a detailed analysis of the step recognition and anticipation branches. Moreover, we conduct an ablation study on the impact of different prompt structures on the performance of the Large Language Model for mistake detection. Lastly, we analyze the split of \epicO. 
The code is available at \href{https://github.com/aleflabo/PREGO}{https://github.com/aleflabo/PREGO}.

\begin{appendix}

\section{Modelling Details} \label{sec:implementation}
In this section, we discuss the step recognition architecture, delve into details on symbolic reasoning, and provide insights into the hyperparameters used.

\subsection{Step Recognition} 
\label{sec:pictorial}

As shown in Table 2 of the main paper, we evaluate PREGO using two methods for the step recognition branch: OadTR~\cite{OadTR} and \cite{An2023MiniROADMR}, adapted for the task of online mistake detection. This section discusses the OadTR architecture, which achieves the best results on the \epicO dataset. 

OadTR comprises a Transformer Encoder with three encoding layers. Each layer incorporates a Multi-Head self-attention module and an Element-Wise addition module. Similarly to~\cite{OadTR}, PREGO retains the learnable task token along with the frame features, which acquire global discriminative features for the online action detection task. 
The OadTR-based step recognition model is selected according to its recognition performance. In Tab.~\ref{tab:recognition}, we compare two alternatives, which we consider: the original encoder-decoder OadTR~\cite{OadTR} model, and an encoder-only variant, which we propose for PREGO. Note that performance is evaluated on the \assemblyO target benchmark, so the reported estimates differ from what reported when evaluated on~\cite{sener2022assembly101}.
The encoder-only model outperforms the original OadTR architecture for all the window lengths, so we selected it for PREGO to yield action recognition online.
The optimal window size, set at 512, demonstrates a 10\% improvement over OadTR. \
Further to performance, the encoder-only model also significantly reduces the parameter count compared to the original model, reducing it approximately three times in size.

In Table~\ref{tab:recognition}, for the encoder-decoder (OadTR) and encoder-only (PREGO) models, we compare performances achieved by varying the window size, i.e. varying the length in frames of the ingested video excerpt, as input for the recognition model.
The mAP exhibits an ascending trend with increasing window sizes, reaching its maximum when the window size equals 512. Beyond this point, the mAP decreases, suggesting that there is saturation and that the model fails to handle the long-term dependencies between frames.

The inference Runtime, computed on a single sample, shows the advantage of using the PREGO architecture due to the strict time requirements of the online setup. The chosen architecture is approximately 45\% faster than the original model.

\begin{table}[t]

\centering{

\caption{
mAP performance of OadTR~\cite{OadTR} as the Step Recognition method considering different window sizes and architectures on \assemblyO. 
}

\label{tab:recognition}
\resizebox{\linewidth}{!}{
\begin{tabular}{l|cc|ccccc} 
\hline
& \textbf{Parms.} & \textbf{Runtime } & \multicolumn{5}{c}{\textbf{Window size}} \\ 
& & ({\it sec}) &\textbf{64} & \textbf{128} & \textbf{256} & \textbf{512} & \textbf{768} \\ 

\hline\hline

Encoder-Decoder (\textit{OadTR}) & 74 M & 0.031 & 11.0 & 12.1 & 12.7 & 13.0 & 12.2 \\ 
Encoder \textit{(PREGO)} & 21 M  & 0.017  & 11.3 & 12.3 & 12.8 & \textbf{14.5} & 13.2 \\

\hline
\end{tabular}
}
}

\end{table}

\begin{figure*}[!t]
    \centering
    \includegraphics[trim={0.5cm 2cm 0.5cm 2cm}, clip, width=0.9\linewidth]{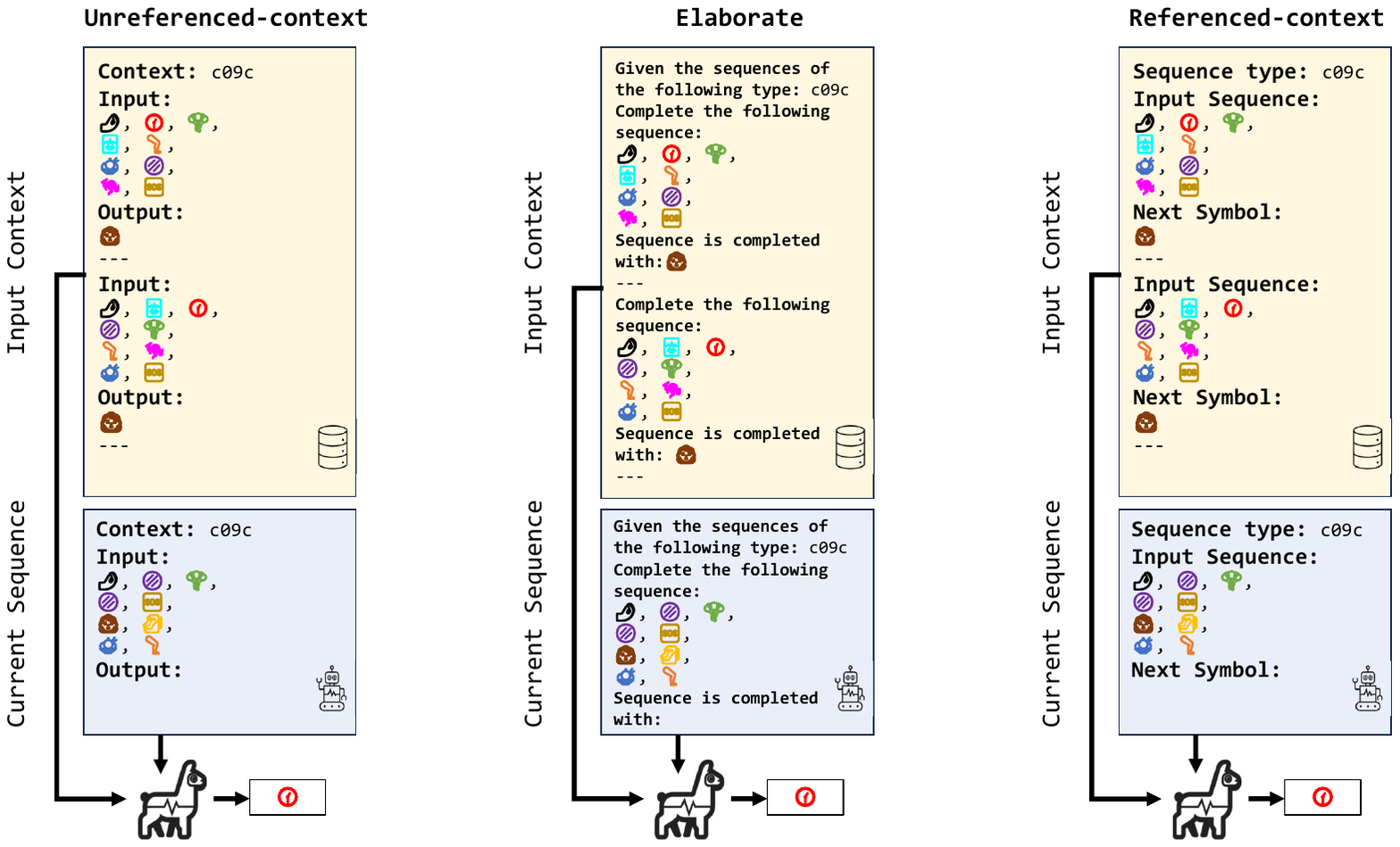}
    \caption{
    Three different variants, defining different inputs to the LLMs.
    On the left, the prompt lacks any reference to sequences or symbols to be completed. In the center, the prompt consists of detailed and lengthier requests. 
    On the right, the prompt incorporates the context of the sequence explicitly. This third variant performs best and it is therefore adopted in PREGO.
    }
    \label{fig:ablation}
\end{figure*}

\subsection{Symbolic Reasoning}

The main paper shows two distinct setups for 
the step anticipation branch of PREGO, 
i.e., one using Llama-2 and one adopting the OpenAI GPT-3.5.
Specifically, we employ the 7 billion parameters version of Llama-2~\cite{touvron2023llama} as the LLM module for symbolic reasoning.
We adjust the temperature and output tokens hyperparameters to 0.6 for the former, which aims to enforce quasi-deterministic outputs, while setting the latter to 4 to ensure answers of the desired length.

The second selected LLM is GPT-3.5~\cite{ouyang2022training}, developed by OpenAI, provided as a paid API service. 
It has been trained with reinforcement learning employing reward models learned by human feedback~\cite{ouyang2022training}. 
In our experiments, we fix the temperature to 0.0.
Unlike Llama-2, we do not constrain this model on the output length. This is because, in our experiments, 
GPT-3.5 was more likely to give answers consistent with the form of the prompt when compared with Llama-2.

\section{Prompt Context}
\label{sec:prompt}
In Sec. 5.4 of the main paper, we discussed the performance of the Step Anticipation branch using different prompts. In Fig.~\ref{fig:ablation}, we show the three prompts, dubbed "Unreferenced-context", "Elaborate", and "Referenced-context". As reported in the main paper, the results of PREGO with the three versions are similar, hence the symbolic reasoning of the LLM is not affected by the input prompts.

\begin{figure}[tbp!]
    \centering
    \includegraphics[trim={2.5cm 1.8cm 2.5cm 2.5cm}, clip, width=\linewidth]{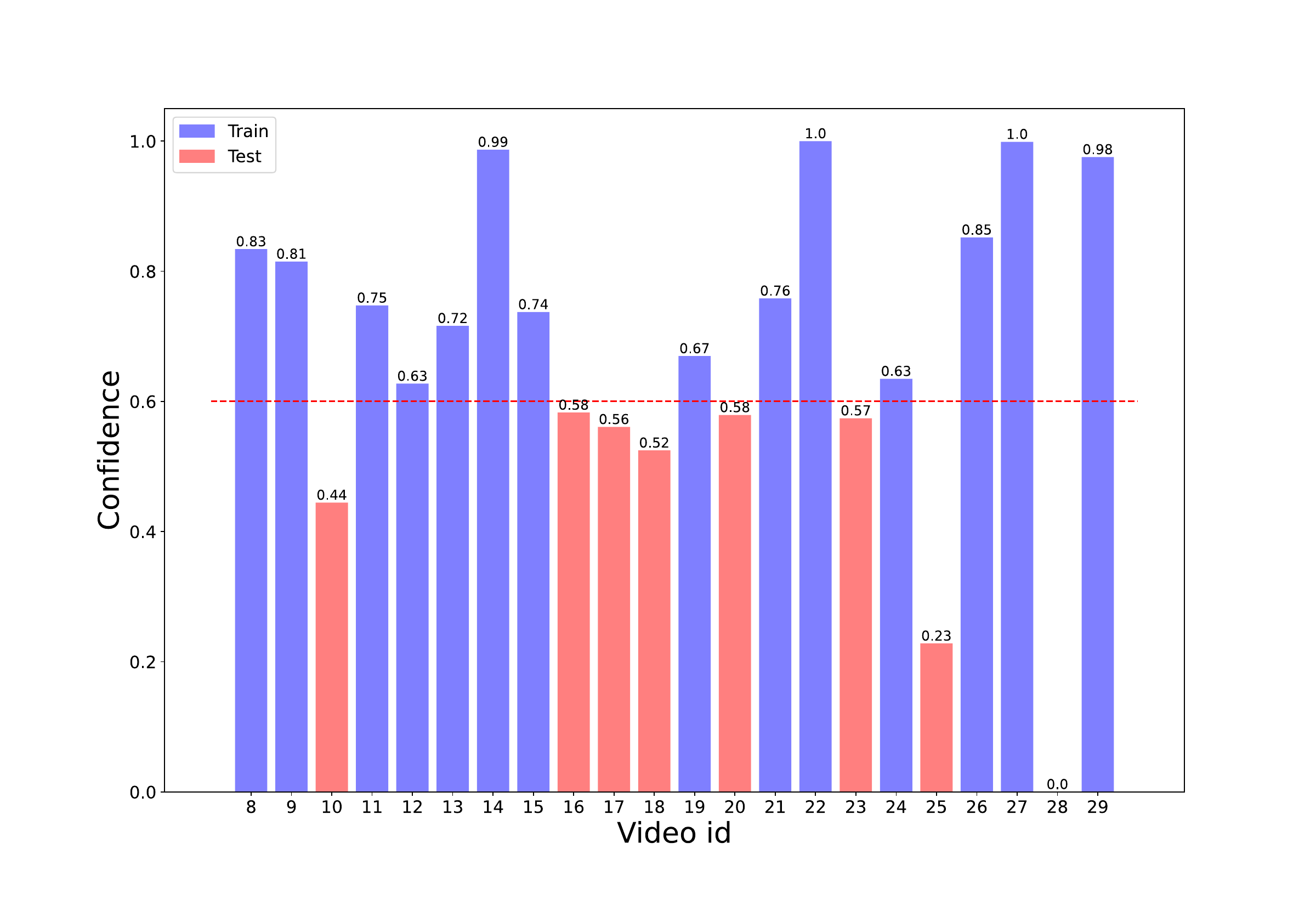}
    \caption{\epicO split between train and test set based on the self-confidence of actors while performing the procedure. The videos with id between $[1,7]$ do not have confidence score annotations and are included in the test set.   
    }
    \label{fig:uncertainty}
\end{figure}

\section{\epicO split} \label{sec:epico}
As described in Sec. 3.1.2 of the main paper, we propose a new split of the \epic dataset. We opt to include this dataset since it includes different types of procedural mistakes, i.e. ordering, omissions, repetitions, and corrections. It has been recorded open-air, differentiating it from the assembly and kitchen-based datasets in literature. 
\epic consists of 29 videos, all of which contain annotated procedural errors. To follow the OCC paradigm, we split the dataset according to the confidence the actors annotated while performing the procedure, as shown in Fig.~\ref{fig:uncertainty}. The videos that form the test set have a median confidence score under 0.6, while the others form the train set. Only 22 videos have the confidence score annotations, while the remaining 7 do not and are assigned to the test set.

\end{appendix}

\end{document}